\documentclass{llncs}

\usepackage[T1]{fontenc}
\usepackage{graphicx}
\usepackage{booktabs}
\usepackage{multirow}
\usepackage{silence}
\usepackage{amsmath}
\usepackage{xcolor}
\usepackage{hyperref}

\let\oldthebibliography\thebibliography
\renewcommand{\thebibliography}[1]{%
  \oldthebibliography{#1}%
  \setlength{\itemsep}{0pt}\setlength{\parsep}{0pt}\setlength{\parskip}{0pt}}

\makeatletter
\renewcommand\section{\@startsection{section}{1}{\z@}%
   {-12\p@ \@plus -3\p@ \@minus -3\p@}%
   {8\p@ \@plus 3\p@ \@minus 3\p@}%
   {\normalfont\large\bfseries\boldmath
    \rightskip=\z@ \@plus 8em\pretolerance=10000 }}
\renewcommand\subsection{\@startsection{subsection}{2}{\z@}%
   {-12\p@ \@plus -3\p@ \@minus -3\p@}%
   {6\p@ \@plus 3\p@ \@minus 3\p@}%
   {\normalfont\normalsize\bfseries\boldmath
    \rightskip=\z@ \@plus 8em\pretolerance=10000 }}
\makeatother

\usepackage{float}

\begin{document}

\title{Interpretable Patch-Based Deep Learning for Wildfire Spread Prediction from Ensemble Simulations}

\titlerunning{Deep Learning Surrogates for Wildfire Spread Prediction}

\author{
     Marcin Lawenda\inst{1} 
\and Aleksandra Krasicka\inst{1} 
\and David Caballero\inst{2}
\and Luis Torres\inst{2}
\and Łukasz Szustak\inst{3}
}
\authorrunning{Marcin Lawenda et al.}

\institute{
Poznan Supercomputing and Networking Center, Poznań, Poland \\
\email{lawenda@man.poznan.pl, akrasicka@man.poznan.pl} 
\and
MeteoGrid, Madrid, Spain \\
\email{david@meteogrid.com, luis@meteogrid.com}
\and
Technical University of Czestochowa, Częstochowa, Poland \\
\email{lukasz.szustak@pcz.pl}
}

\maketitle

\begin{abstract}
Wildfire spread is traditionally predicted using physics-based simulators, which are physically interpretable but whose cost increases with each additional ensemble member. We ask how well deep learning surrogates can reproduce these simulations at a fraction of this cost, training them on 10{,}584 fire spread simulations at 2~m resolution for the Rectoret region in Catalonia, Spain. Four architectures are compared: a patch-based U-Net, a transfer-learned ResNet-50, a physics-informed network constrained by the wind-driven advection equation and a Swin-Unet transformer. Among the terrain and vegetation variables, only surface fuel load predicts burn probability with any strength ($r = 0.27$) and including it lowers prediction error by 21\%. The remaining variables correlate weakly and are highly duplicative. Next, an experiment with saliency, occlusion and rotation demonstrates the models' learning. Convolutional models rely primarily on distance from the current fire front, while Swin-Unet assigns more weight to fuel and terrain, a finding also noted in an unrelated wildfire dataset. When applied without retraining to the second region, Pedriza, all three convolutional models still predict fire spread, losing accuracy by a small but systematic margin.

\end{abstract}

\section{Introduction}
\label{sec:intro}

Wildfires are among the most devastating natural hazards affecting Mediterranean landscapes, and their increasing frequency and severity are linked to climate change and land-use changes in the wildland-urban interface. Accurate, spatially precise fire spread prediction influences evacuation planning, firefighting resource allocation and fuel-break design. Traditional approaches rely on physically based or semi-empirical simulators, particularly cellular automata (CA) models based on Rothermel's fire-spread equations \cite{rothermel1972} (e.g.\ FARSITE \cite{finney1998farsite}, FlamMap \cite{finney2006flammap}). These simulators are physically interpretable and well-validated. However, their cost increases directly with the number of simulations. A burn probability (BP) map is generated from multiple simulations, each representing a separate, complete run with a new combination of wind and ignition point. Producing a reliable BP map therefore multiplies the cost of a single run many times over.

This cost motivates machine learning surrogates: once trained, the surrogate reproduces the simulator's output in a single forward pass, making inference virtually free. Whether it also performs outside the terrain and fuel on which it was trained is a separate and much more difficult question, which we treat here only as a preliminary comparison (Section~\ref{sec:comparison}). Our dataset consists of a set of fire spread simulations for the Rectoret region in Catalonia, recording the minute the fire reaches each cell under varying wind and ignition conditions, along with the terrain and fuel descriptors that shape it.

Our contribution has three parts, moving from what the data itself shows, to how closely the surrogates reproduce it, to what they have learned in doing so. First, we characterise the statistical relationship between terrain/fuel features and burn probability using correlation, Moran's~I spatial autocorrelation and XGBoost feature importance. Second, we design and compare four deep learning-based models to predict fire spread 30 minutes in advance on $128\times128$ patches: a custom U-Net (108-configuration architecture/loss search), a transfer-learned ResNet-50, a physics-informed variant penalising violations of a wind-driven advection equation and a Swin-Unet transformer baseline. Third, we apply complementary interpretability techniques (saliency, occlusion, local perturbations) to characterise what each model has learned and compare performance on a second region (Pedriza).

\section{Related Work}
\label{sec:related}

The work related to this paper spans two traditions: physics-based fire propagation simulation, which provides the ground truth used here, and trained surrogates, which approximate it at a fraction of the inference cost. We discuss four streams in turn: physics-based simulators, which solve the fire-spread equations, CNN/U-Net surrogates, physics-informed networks, which instead learn from data under a penalty for violating them and transformer-based architectures. Each of these corresponds to one of the models compared in Section~\ref{sec:methods}. We conclude by contrasting our own contributions with them.

\textbf{Physics-based simulators.} Operational fire propagation prediction has historically relied on semi-empirical models based on Rothermel's equations \cite{rothermel1972}, which underpin simulators such as FARSITE \cite{finney1998farsite} and FlamMap \cite{finney2006flammap}, which propagate the fire front by expanding the Huygens wavelet. CA variants discretise the landscape into a grid and propagate the ignition state between neighbouring cells, offering a more parallelisable alternative. Recent, differentiable, GPU-accelerated CA simulators achieve millisecond-per-step computations and can still be calibrated using gradient descent. They also demonstrate better inter-region transfer than purely supervised surrogates \cite{xia2025pytorchfire}. This is a useful warning for Section~\ref{sec:comparison}. The burn probability (BP) formula used here is the fraction of the ensemble in which a cell burns. This is a standard CA simulator output, widely used for fuel management and risk zoning.

\textbf{CNN/U-Net surrogates.} The \emph{Next Day Wildfire Spread} dataset and its CNN baseline \cite{huot2022nextday} established the standard approach to image-to-image segmentation based on stacked terrain, weather and fire status channels, which was later refined in two-branch CNNs by processing fuel and weather separately \cite{han2026firesensenet}. Closer to our setup, ConvLSTM models were trained directly on simulated wildfire data from the mathematical analog model \cite{burge2020convlstm}, setting a precedent for surrogate evaluation based on simulation results rather than observed fires. Transfer learning from backbone networks pretrained on ImageNet \cite{he2016resnet} forms the basis for our ResNet-50 baseline model, and patch-based training handles $1500\times1500$-pixel rasters, from which we assemble 11-channel input tensors.

\textbf{Physics-informed neural networks.} PINNs add differential equations to the training loss, guiding the model toward physically consistent predictions, even with limited labelled data. Wind-driven fire front propagation naturally fits this model, as an advection equation for the fire state field. The closest approximation to our model is \cite{vogiatzoglou2025pinn}, which adds mass and energy conservation constraints to learn forest fire spread parameters from both simulated and real thermal data, and shows that training with physical constraints allows the recovery of significant parameters even from noisy observations. This motivates the advection penalty in Section~\ref{sec:methods}.

\textbf{Transformer-based architectures.} Vision-transformer architectures have recently been applied to forest fire prediction as an alternative to purely convolutional designs, including Swin-based encoder-decoder variants for next-day spread, whose accuracy depends significantly on weights pre-trained on ImageNet \cite{lahrichi2026wsts}. The most relevant element of our Swin-Unet comparison is \cite{zhou2025comparative}, which compares Autoencoder, ResNet, U-Net and Swin-Unet using a decade of California remote sensing data, with Swin-Unet and U-Net outperforming the other two. Grad-CAM analysis also showed that Swin-Unet assigns a higher weight to vegetation and drought than U-Net. We observe the same difference in Section~\ref{sec:results}.

This paper is unique in combining these threads. A data-driven CNN, a transfer-based framework, a physics-informed variant and a transformer are compared within a single dataset and a single patching scheme. Interpretability is examined using three separate methods and a direct comparison is performed across two regions. Most of the papers cited above examine a single model family or a single region, but rarely both simultaneously.

\section{Study Area and Data}
\label{sec:data}

\textbf{Simulation ensemble.} The dataset comprises $NS=10{,}584$ stochastic fire-spread simulations for the Rectoret region, Catalonia, Spain (ETRS89 UTM31N), at 2~m resolution, bounded by \mbox{41\textdegree 24'48.09''N}, \mbox{2\textdegree 4'44.72''E} (lower left) and \mbox{41\textdegree 26'26.69''N}, \mbox{2\textdegree 6'52.50''E} (upper right). Each simulation is a \texttt{TA<Number>.asc} raster encoding the simulated minute of fire arrival at each cell (capped at 180 minutes), parameterised by wind speed (WS), wind direction (WD) and ignition offset (OX, OY). A companion Burn Probability raster summarises the ensemble as
\begin{equation}
BP = 100 \cdot \frac{NF}{NS},
\label{eq:bp}
\end{equation}
where $NF$ is the number of simulations in which fire passed through a given cell.

The four generating parameters were sampled on a \emph{regular full-factorial grid} (Table~\ref{tab:sampling}): three wind speeds, eight wind directions at 45\textdegree increments and a $21\times21$ grid of ignition points spaced 100~m apart across 500--2500~m in each direction. That gives $3\times8=24$ distinct wind configurations, each realised at all $21^2=441$ ignition points, for $24\times441=10{,}584$ simulations in total. The design is therefore exactly balanced, since every wind configuration is represented by the same ignition points and every ignition point by the same winds.

\begin{table}[ht]
\centering
\small
\caption{Sampling of the four ensemble-generating parameters. The design is a full factorial grid, giving $3\times8\times21\times21=10{,}584$ simulations.}
\label{tab:sampling}
\begin{tabular}{llc}
\toprule
Parameter & Values & Count \\
\midrule
Wind speed (WS) & 5, 10, 20~km\,h$^{-1}$ & 3 \\
Wind direction (WD) & 0\textdegree, 45\textdegree, \ldots, 315\textdegree\ (45\textdegree\ steps) & 8 \\
Ignition offset OX & 500, 600, \ldots, 2500~m (100~m steps) & 21 \\
Ignition offset OY & 500, 600, \ldots, 2500~m (100~m steps) & 21 \\
\midrule
\multicolumn{2}{l}{Total ensemble size $NS$} & 10{,}584 \\
\bottomrule
\end{tabular}
\end{table}

\textbf{Terrain and fuel covariates.} The terrain and fuel descriptors come from the Institut Cartogr\`{a}fic de Catalunya in ArcGIS ASCII format. They comprise slope and aspect, a continuous fuel-load map (\texttt{cmb01}) with its binary counterpart (\texttt{cmb}), five vegetation-continuity indices computed at several scales from the WUIX index (\texttt{bio3x3n}, \texttt{bio5x5n}, \texttt{cont6m}, \texttt{cont10m}, \texttt{CONT\allowbreak\_NORM\allowbreak\_2\allowbreak\_20}) and finally surface fuel load and crown fuel. Surface fuel load later proves to be the single most informative covariate for BP prediction (Section~\ref{sec:exploratory}). It is obtained by translating each BEHAVE-Anderson fuel class \cite{anderson1982aids} into a total fuel load in kg\,m$^{-2}$, which for the 2~m grid corresponds to four times the per-square-metre value. In the deep-learning pipeline this raster (\texttt{fuel\_load}) is one of the eleven input channels, min--max normalised to $[0,1]$ along with the other static layers. Crown fuel is not used as a model input (Section~\ref{sec:methods}). The fuel-load raster is \emph{derived from the same fuel-model classification the simulator uses to propagate fire}, so it is not an independent predictor. 

\textbf{Second region: Pedriza.} For a cross-region comparison, a second region was used: La Pedriza de Manzanares, Community of Madrid, Spain (ETRS89 UTM30N, EPSG:25830), bounded by \mbox{40\textdegree 42'21.13''N}, \mbox{3\textdegree 57'27.98''W} (lower left) and \mbox{40\textdegree 48'6.63''N}, \mbox{3\textdegree 48'54.53''W} (upper right). It is centred on a granite massif in the southern Sierra de Guadarrama rising from about 890~m to 2{,}029~m, with Mediterranean vegetation of pine, holm oak, juniper and scrubland. The raster is $1217\times1054$ cells at 10~m resolution, so the domain spans 12.17~km east--west by 10.54~km north--south, about 128~km$^2$, roughly fourteen times the area of the Rectoret domain. Pedriza therefore shares the broad fuel and terrain regime of Rectoret while differing in elevation, geology and in covering a much larger and more varied area. One difference matters for everything that follows. Rectoret is mapped at 2~m and Pedriza at 10~m, so a $128\times128$ patch covers 256~m in Rectoret but 1280~m in Pedriza, twenty-five times the ground area. The region was used in two ways: descriptively, aggregated to a 100~m/200~m mesh with per-cell mean, standard deviation, minimum and maximum for slope, aspect, bio5x5n, cont10m, CONT\_NORM\_2\_20, fuel load and BP, and as a model-transfer test bed at native patch resolution.

\section{Exploratory Analysis}
\label{sec:exploratory}

\begin{figure}[ht]
\centering
\includegraphics[width=0.8\textwidth]{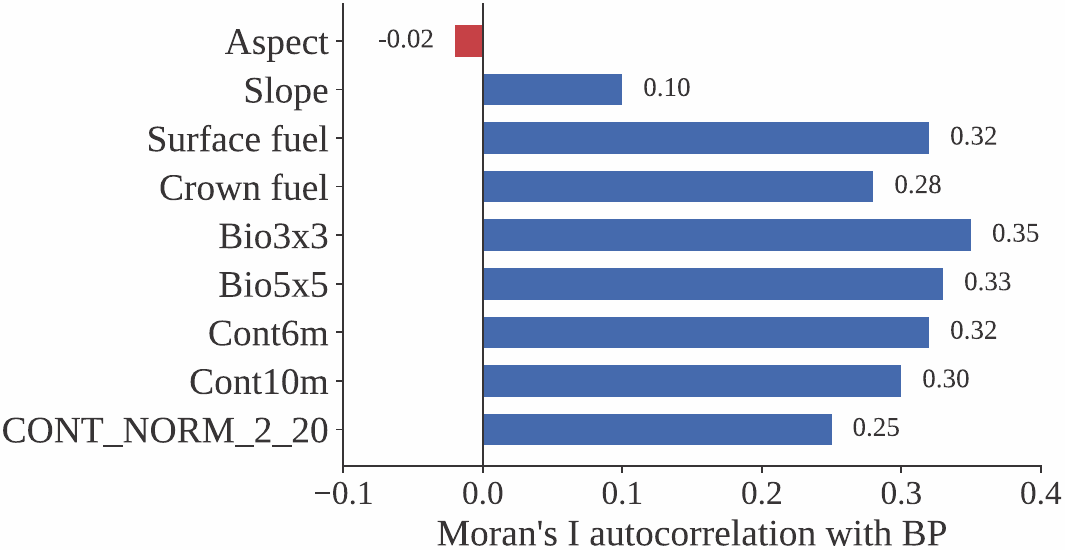}
\caption{Moran's~I spatial autocorrelation with Burn Probability for each terrain and fuel covariate. The vegetation-continuity and fuel descriptors cluster moderately, while slope and aspect are close to zero.}
\label{fig:morans}
\end{figure}

\textbf{Correlation and spatial autocorrelation.} Both BP and the vegetation-related features are moderately clustered in space (Moran's~I \cite{moran1950}), with the vegetation-continuity and fuel descriptors all falling between $I\approx0.25$ and $0.35$: bio3x3 $0.35$, bio5x5 $0.33$, surface fuel $0.32$, cont6m $0.32$, cont10m $0.30$, crown fuel $0.28$ and CONT\_NORM\_2\_20 $0.25$. Slope ($0.10$) and aspect ($-0.02$) show almost none at this scale (Fig.~\ref{fig:morans}).

\begin{figure}[ht]
\centering
\includegraphics[width=0.7\textwidth]{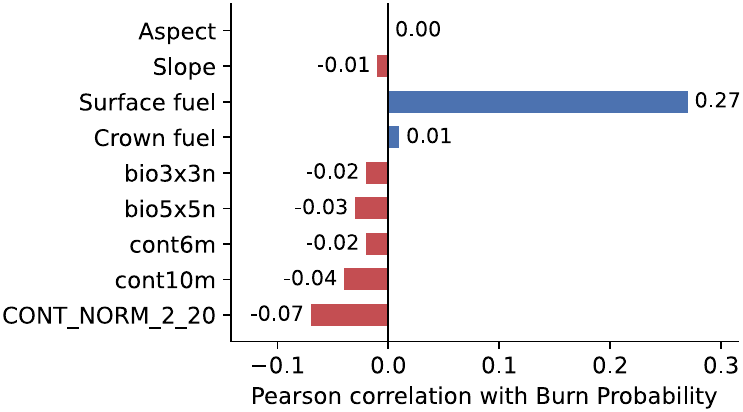}
\caption{Pearson correlation of each terrain and fuel covariate with Burn Probability, at the pixel level. Surface fuel load is a clear outlier.}
\label{fig:corr-bp}
\end{figure}

 Correlation with BP tells a similar story: at the pixel level nothing correlates strongly on its own, and aspect, slope and the continuity descriptors all sit at $|r|\le0.07$. The one exception is \textbf{surface fuel load} ($r=0.27$), which is why we single it out below (Fig.~\ref{fig:corr-bp}).

 At the coarser aggregated-cell level, the vegetation-continuity descriptors correlate strongly with each other, mostly above $r=0.95$. They therefore carry largely the same information. The correlation between \texttt{aspect\_max} and the continuity and biomass maxima is more moderate, at $r\approx0.73$--$0.83$.

\begin{figure}[ht]
\centering
\includegraphics[width=0.5\textwidth]{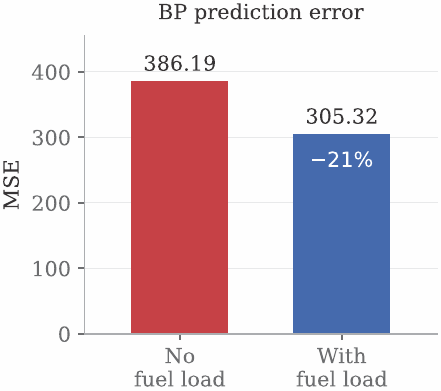}
\caption{BP prediction MSE with and without surface fuel load. Including it reduces MSE from 386.19 to 305.32, a 21\% reduction.}
\label{fig:feat-imp-mse}
\end{figure}

\textbf{Feature importance via gradient boosting.} We trained an XGBoost \cite{chen2016xgboost} model to predict BP from the terrain and fuel covariates, once with surface fuel load and once without it. Including surface fuel load reduces the mean squared error (MSE) from \textbf{386.19 to 305.32}, a 21\% reduction. It therefore improves the prediction, rather than merely correlating with BP by coincidence (Fig.~\ref{fig:feat-imp-mse}).

\begin{figure}[ht]
\centering
\includegraphics[width=0.7\textwidth]{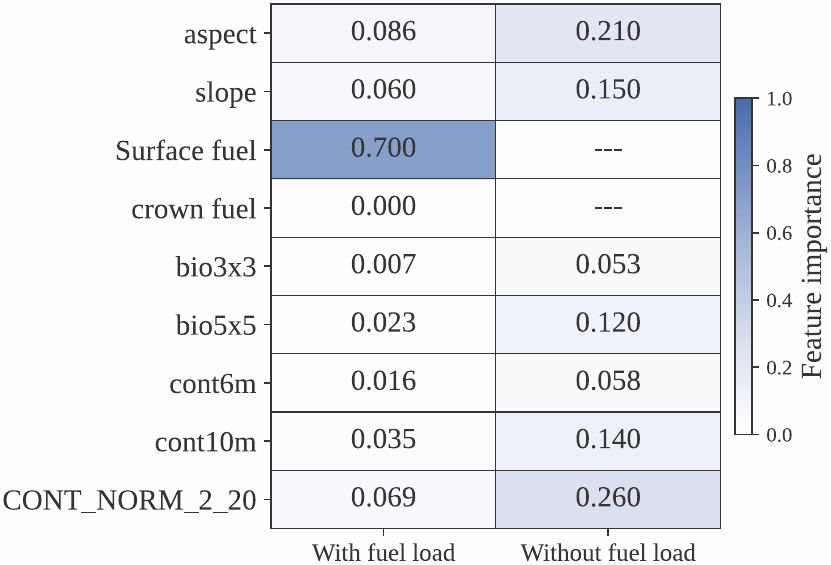}
\caption{XGBoost feature importance for BP prediction, with and without surface fuel load. Surface fuel load dominates when present (0.700). Removing it redistributes importance to CONT\_NORM\_2\_20 and aspect. Crown fuel contributes essentially nothing.}
\label{fig:feat-imp-grid}
\end{figure}

 With surface fuel load included it absorbs almost all of the model's importance (0.700), leaving the remaining covariates marginal (CONT\_NORM\_2\_20 0.069, aspect 0.086, cont10m 0.035, bio5x5n 0.023, cont6m 0.016, bio3x3 0.007) and crown fuel at effectively zero (0.000). Removing it redistributes importance towards CONT\_NORM\_2\_20 (0.260), aspect (0.210), slope (0.150) and cont10m (0.140), (Fig.~\ref{fig:feat-imp-grid}).

\textbf{Wind sensitivity.} Because wind speed and direction are constant within a simulation, they cannot enter the pixel-level analysis directly. Grouping simulations instead by direction at fixed speed and by speed at fixed direction (180\textdegree), shows the effect: BP rises markedly for wind from 315\textdegree, and higher speeds enlarge and merge the high-BP regions. This sits awkwardly beside the small importance the models assign to wind (Section~\ref{sec:results}), a tension we return to in Section~\ref{sec:discussion}.

\section{Methods}
\label{sec:methods}

\textbf{Problem formulation.} We frame fire-spread prediction as image-to-image translation: given a $128\times128$ patch describing the current fire state, the terrain and fuel covariates and the wind, the model predicts the binary fire state 30 minutes later. Patches are extracted from the $1500\times1500$-pixel simulation domain with a 64-pixel stride, so neighbouring training patches overlap by half. That domain holds the single-channel arrival-time raster, which we combine with the static terrain and fuel layers and with the wind. Eleven channels make up the input. One carries the current fire state as a binary mask and two more carry the distance to the fire front, at the current time and 30 minutes earlier. Six hold the static terrain and fuel layers, namely aspect, slope, fuel load, bio5x5n, cont10m and CONT\_NORM\_2\_20. The remaining two carry wind speed and direction, each broadcast across the patch as a constant matrix. Patches with $<1\%$ fire-active pixels are discarded, with no-spread patches retained at only 0.1\% probability to limit class imbalance.

\textbf{Custom U-Net.} We selected the structure and hyperparameters by grid search on a 1{,}000-simulation subset. The search covered 3 activation functions (ReLU, Swish, Leaky ReLU), 6 loss functions and 6 filter-depth configurations, giving 108 variants trained for 10 epochs at batch size 10. The loss functions were binary cross-entropy (BCE), Matthews correlation \cite{matthews1975,chicco2020mcc}, Dice \cite{dice1945,milletari2016vnet}, Hausdorff \cite{karimi2019hausdorff}, Tversky \cite{salehi2017tversky} and Focal Tversky \cite{abraham2019focal}. All variants follow a U-Net \cite{ronneberger2015unet} encoder--bottleneck--decoder structure with a final $128\times128$ sigmoid output. Ranked first was Leaky ReLU with base filter width 64, depth 2 and a combined $0.5\,$BCE$\,+\,0.5\,$Dice loss. The configuration carried forward to all later experiments, named ``Custom'' in the tables and figures below, is the same one at base filter width 32 (depth 2, $[32,64]$), which ranked third and differs only marginally on the combined score. Shallower networks and Dice-based losses generally did better, which we attribute to the strong class imbalance in fire segmentation, where burned pixels are rare. The BCE--Dice combination is an established choice for such tasks \cite{shahid2023ffsunet}.

\textbf{ResNet-50 baseline.} We adapted an ImageNet-pretrained ResNet-50 \cite{he2016resnet} (\texttt{include\allowbreak\_top=\allowbreak False}, $128\times128\times3$ input). A \textbf{learned $1\times1$ convolution} (the \texttt{input\_projection} layer) first projects the 11 input channels onto 3. A decoder then restores the $128\times128$ output. It has five nearest-neighbour $2\times$ upsampling stages, each followed by a $3\times3$ ReLU convolution with $512\to256\to128\to64\to32\to16$ filters and ends with a $1\times1$ sigmoid convolution. Unlike the Custom U-Net, the decoder has \textbf{no encoder--decoder skip connections}. The backbone stays frozen for the first 40 epochs and is then unfrozen for fine-tuning, at which point the Adam \cite{kingma2015adam} learning rate drops from $10^{-4}$ to $10^{-5}$. Activation is ReLU throughout, with the same $0.5\,$BCE$\,+\,0.5\,$Dice loss used for Custom.

\textbf{Physics-informed neural network (PINN).} The PINN shares the Custom architecture but augments the loss with a physics-inspired penalty encoding wind-driven advective transport of the fire-state field $u(x,y,t)$. With $\theta=2\pi\cdot WD$, wind velocity components are
\begin{align}
v_x = WS\cos\theta, \qquad v_y = WS\sin\theta,
\end{align}
and spatial gradients are approximated by finite differences. The governing advection equation $\partial u/\partial t + v_x\,\partial u/\partial x + v_y\,\partial u/\partial y = 0$ is enforced via
\begin{equation}
L = L_{data} + \lambda_1 L_{smooth} + \lambda_2 L_{adv}, \quad L_{adv} = \left(v_x \tfrac{\partial u}{\partial x} + v_y \tfrac{\partial u}{\partial y}\right)^2,
\label{eq:pinn-loss}
\end{equation}
combined with a spatial-smoothness penalty $L_{smooth}$ and a mean-squared-error data-fitting term $L_{data}$. In the released implementation $\lambda_1=0.05$ weights the smoothness term and $\lambda_2=0.1$ the advection term. In every other respect the PINN matches the Custom U-Net (\texttt{base\_filters=32}, depth~2, Leaky ReLU), trained for 100 epochs at batch size 10 through a custom training step. A sweep over $\lambda_2\in\{0,0.05,0.1,0.2\}$ across three seeds leaves the test Dice loss between 0.086 and 0.090, with no significant effect (Friedman, $p=0.46$) and seed-to-seed variation larger than the spread between settings.

\textbf{Swin-Unet.} The fourth model replaces the convolutional encoder with a transformer one. We use SwinTransformerV2-Tiny \cite{liu2021swin} with window size 8, pretrained on ImageNet: four stages of $[2,2,6,2]$ blocks with $[3,6,12,24]$ attention heads, embedding dimension 96 and a stem patch size of 4. Apart from the backbone it is built exactly like the ResNet-50 baseline, sharing the same learned $1\times1$ channel projection, the same decoder of five upsampling stages, the same absence of skip connections and the same 40-epoch backbone freeze. Since the two differ only in their encoder, the comparison between them is a controlled one. Following earlier transformer-based wildfire work \cite{lahrichi2026wsts,zhou2025comparative}, we include it to test what changes when the encoder attends globally, both in accuracy and in what the network learns. This is a Swin encoder with a convolutional decoder rather than the symmetric transformer decoder of the original Swin-Unet. We keep the name ``Swin-Unet'' for consistency with the figures.

\section{Experimental Setup}
\label{sec:experimental-setup}

Training used mini-batches of size 10 per epoch, with structure/hyperparameters fixed per Section~\ref{sec:methods} and applied consistently across models. The ensemble is split \emph{by simulation}. A simulation goes into the test set if its ignition offset $(OX,OY)$ falls on a 200~m grid, which represents every wind configuration equally. All remaining simulations form the training set. This gives $10\times10=100$ test ignition points per wind configuration, so $24\times100=2{,}400$ test simulations (22.7\%) and $8{,}184$ training simulations (77.3\%). Patches are extracted within each set separately, so no simulation contributes patches to both. At prediction time, patches are re-assembled into a full-domain map by averaging overlapping predictions, then thresholded at $0.5$ to a binary classification. We report five metrics: Mean Absolute Error (MAE), Mean Squared Error (MSE), Dice loss, the Matthews correlation coefficient (MCC) \cite{matthews1975,chicco2020mcc} and a custom weighted error. Dice loss is overlap-based and suited to class imbalance. MCC summarises the whole confusion matrix as a single correlation between prediction and truth, from $-1$ to $+1$, and stays informative under class imbalance. The weighted error is
\begin{equation}
\text{Error} = \frac{w_{TP}\,TP + w_{TN}\,TN - w_{FP}\,FP - w_{FN}\,FN}{TP + TN + FP + FN},
\label{eq:weighted-error}
\end{equation}
where $TP$, $TN$, $FP$ and $FN$ are pixel counts of true and false positives and negatives. The four weights, which let false positives and false negatives be penalised differently, are configurable. All values reported here use one fixed setting, so weighted errors are comparable across models but not on an absolute scale. Evaluation is stratified by \textbf{fire size}, with simulations grouped into four bins by final burned-area fraction (0--0.54\%, 0.55--1.11\%, 1.12--1.62\% and 1.70--2.24\%).

\section{Results}
\label{sec:results}

\subsection{Quantitative model comparison}

Custom and PINN are effectively tied at the front. A paired Wilcoxon signed-rank test over the 16 shared evaluation runs finds no significant difference between them ($W=64$, $p=0.86$; mean weighted error $0.230$ against $0.230$). The PINN is the Custom U-Net with the physics terms added to its loss and nothing else changed, so the penalty does not measurably improve accuracy at this scale. The two also use $0.47$~M parameters against $34.6$~M for ResNet-50 and $32.7$~M for Swin-Unet, so roughly seventy times fewer parameters give better accuracy. They lead on every metric and train in the least time (Table~\ref{tab:results}). ResNet-50 is behind (MCC $0.767$) and Swin-Unet further still (MCC $0.499$), despite both being the slowest to train. The weighted error gives the same ordering: $0.277$ for Custom, $0.284$ for PINN, $0.861$ for ResNet-50 and $2.353$ for Swin-Unet, averaged over the four fire-size groups at fire stages 30--120~min with a 25\% patch step. Error is somewhat higher for the smallest and largest fire-size groups than for intermediate ones.

\begin{table}[ht]
\centering
\small
\caption{Model comparison on 1{,}500 held-out patches, all four models evaluated on the same data. Training time is for 100 epochs on the GPU used for the original runs.}
\label{tab:results}
\begin{tabular}{lcccccc}
\toprule
Model & Params & MAE~$\downarrow$ & MSE~$\downarrow$ & Dice~$\downarrow$ & MCC~$\uparrow$ & Train (h) \\
\midrule
Custom (U-Net) & \textbf{0.47\,M} & \textbf{0.024} & 0.019 & \textbf{0.043} & 0.945 & \textbf{5.8} \\
ResNet-50      & 34.6\,M & 0.104 & 0.078 & 0.187 & 0.767 & 7.5 \\
PINN           & \textbf{0.47\,M} & 0.031 & \textbf{0.018} & 0.057 & \textbf{0.946} & 5.9 \\
Swin-Unet      & 32.7\,M & 0.216 & 0.194 & 0.361 & 0.499 & 8.7 \\
\bottomrule
\end{tabular}
\end{table}

\subsection{Interpretability analysis}

\begin{figure}[ht]
\centering
\includegraphics[width=1\textwidth]{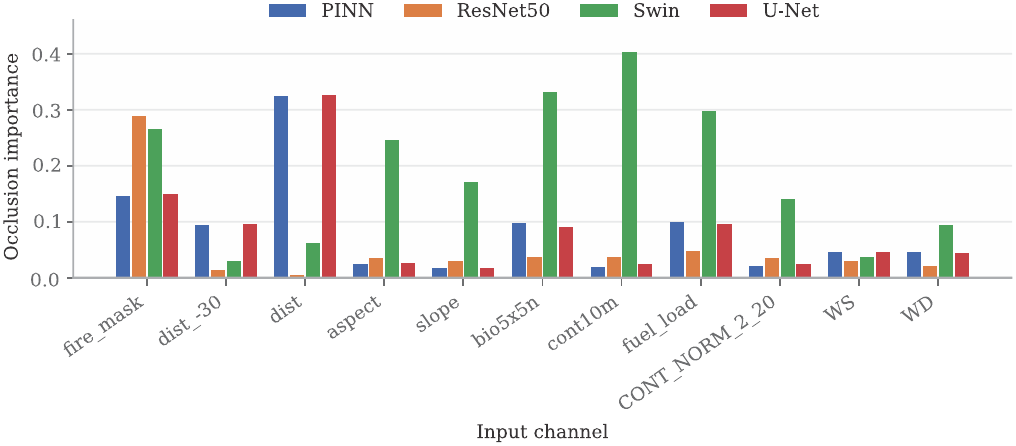}
\caption{Occlusion importance by model across all eleven input channels, measured as the mean change in predicted fire state when each channel is zeroed. Computed on 200 held-out patches. \texttt{dist\_-30} and \texttt{dist} are the distances to the fire front 30 minutes earlier and at the current time.}
\label{fig:interp-model}
\end{figure}

Gradient-based saliency and occlusion sensitivity give a clear ranking for Custom and PINN. The distance to the current fire front dominates (occlusion importance $0.326$ and $0.324$), followed by the current fire mask ($0.149$, $0.146$), then fuel load and bio5x5n at around $0.09$--$0.10$. ResNet-50 behaves differently, relying almost entirely on the raw fire mask ($0.289$, half its total importance) while essentially ignoring the distance channels, for which it records the lowest value anywhere in the analysis ($0.004$ for \texttt{dist}). It never learned to exploit the distance-transform representation the other convolutional models depend on, which may be why it is the weakest of them. An earlier version of this analysis omitted the binary fire mask and including it matters: the mask ranks second for Custom and PINN and \emph{first} for ResNet-50, so its omission had left out the top-ranked channel for one of the four models. Adding Swin-Unet (Fig.~\ref{fig:interp-model}) reveals a split between the two architectures. \textbf{Swin-Unet shifts importance toward fuel and terrain}: its four highest channels are cont10m ($0.403$), bio5x5n ($0.331$), fuel load ($0.299$) and aspect ($0.245$), while the current fire-front distance falls to $0.061$. The convolutional models show the opposite ordering. The transformer therefore appears to learn a different representation, one that relies more on static terrain and fuel. The same effect has been reported on a different wildfire dataset, where Swin-Unet's Grad-CAM attributions also favoured vegetation and drought over a plain U-Net's \cite{zhou2025comparative}. Finding it in two separate datasets and pipelines suggests it is a genuine property of Swin-style attention.

\begin{figure}[ht]
\centering
\includegraphics[width=0.9\textwidth]{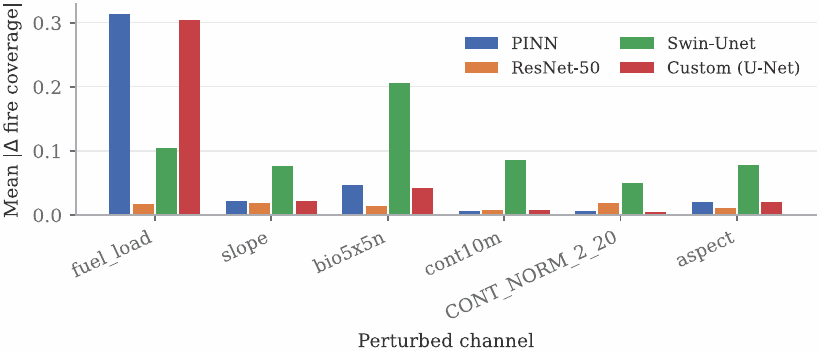}
\caption{Local perturbation along the fire perimeter: mean absolute change in predicted fire coverage when each terrain or fuel channel is perturbed. PINN and the custom U-Net respond overwhelmingly to fuel load, Swin-Unet responds broadly across channels and ResNet-50 is comparatively insensitive to all of them.}
\label{fig:perturb}
\end{figure}

We also perturbed each of the six static terrain and fuel channels locally, along the fire perimeter (Fig.~\ref{fig:perturb}). For the PINN and the custom U-Net, \texttt{fuel\_load} is by far the most influential: perturbing it produces the widest change in fire coverage. Slope produces a smaller response, but one with a consistent sign. The vegetation-continuity descriptors (cont10m, bio5x5n, CONT\_NORM\_2\_20) fall in between. How pronounced this is depends strongly on the model: Swin-Unet responds broadly across all six channels, while ResNet-50 is comparatively insensitive to all of them. Taken together the three methods agree only in part. Perturbation points to fuel load and for Swin-Unet to terrain and fuel more generally, whereas occlusion points to fire-front distance for the convolutional models. Saliency spreads importance more evenly than either. Wind has almost no importance in the saliency and occlusion analyses. We attribute this to how wind is encoded: it is constant across a patch, so methods that look for spatial variation have little to detect. Wind was not part of the perturbation analysis, which covers only the static terrain and fuel channels. To separate the two explanations we rotated the wind direction by a known amount and measured how far the prediction moved, taking the local spread direction from the displacement of the fire front between the two input stages. The result is unambiguous. Every model responds and the response grows monotonically with the rotation: for the Custom U-Net the mean absolute change in predicted fire state is $0.030$ at 45\textdegree, $0.052$ at 90\textdegree and $0.066$ at 180\textdegree, against $0.045$ for simply zeroing the channel. PINN behaves almost identically ($0.069$ at 180\textdegree), Swin-Unet is the most wind-sensitive ($0.123$) and ResNet-50 the least ($0.027$). Reversing the wind changes the prediction most, which is the physically expected ordering.

Wind is therefore \emph{not} ignored and describing its importance as negligible was too strong. What the earlier analysis measured was limited by the encoding: because wind is constant across a patch, methods that look for spatially varying signal understate it. The ranking itself does not change, however. Even a full reversal moves the prediction by about $0.07$ for the convolutional models, against $0.33$ for occluding the fire-front distance, so wind remains a secondary factor behind the current fire state.

\subsection{How much comes from the fire front?}

The distance channels dominate the occlusion ranking, raising the question of whether the surrogate learns environmental drivers or mainly extrapolates the existing front. We retrained the Custom U-Net on reduced inputs under the same protocol, three seeds each (Table~\ref{tab:dist-ablation}). These are interpretability probes, not deployment candidates.

Dropping either distance channel costs almost nothing. Dropping the fire mask as well halves accuracy. The fire mask and the two distance layers are partially redundant representations of the same fire-state information rather than three independent predictors. \texttt{dist} is a distance transform of the fire mask, so removing it leaves the geometry in the mask. Occlusion confirms this, fire-mask importance doubling from $0.121$ to $0.248$ while the eight environmental channels stay flat in total ($0.399$ to $0.342$), the model shifting to another encoding of the same information. The distance channels are causally informative for near-term spread, not shortcuts in the pejorative sense, but their dominance limits what accuracy reveals about environmental sensitivity. Terrain, fuel and wind alone still reach MCC $0.42$, so that signal is real but second-order.

\begin{table}[ht]
\centering
\small
\caption{Input ablation on the Custom U-Net, mean $\pm$ standard deviation over three seeds.
Geometry means the fire mask and both distance channels.}
\label{tab:dist-ablation}
\begin{tabular}{lcccc}
\toprule
Input & Ch. & MSE~$\downarrow$ & Dice~$\downarrow$ & MCC~$\uparrow$ \\
\midrule
Full                        & 11 & $0.025\pm0.001$ & $0.057\pm0.003$ & $\mathbf{0.920}\pm0.005$ \\
No \texttt{dist}            & 10 & $0.029\pm0.001$ & $0.063\pm0.001$ & $0.914\pm0.001$ \\
No \texttt{dist\_-30}       & 10 & $0.026\pm0.003$ & $0.060\pm0.006$ & $0.919\pm0.006$ \\
No distance channels        &  9 & $0.028\pm0.001$ & $0.061\pm0.001$ & $0.917\pm0.002$ \\
No geometry                 &  8 & $0.162\pm0.007$ & $0.402\pm0.015$ & $0.424\pm0.015$ \\
\bottomrule
\end{tabular}
\end{table}

\section{Comparison with a Second Region}
\label{sec:comparison}

Because AI-based fire-spread models are often region-specific and degrade in unseen environments, we applied the trained models to Pedriza as a first check. It differs from Rectoret in terrain and environment, but one difference matters more than the rest. Because Pedriza is mapped at 10~m against Rectoret's 2~m, a patch of the same pixel size covers twenty-five times the ground area (Section~\ref{sec:data}), so the models were trained at one scale and applied at another. Any gap reported below therefore mixes a change of region with a change of spatial scale and the present data cannot separate the two. Setting scale aside, the correlation structure of the aggregated cells is similar in the two regions, so they are comparable in statistical structure even though their absolute values differ. Quantitatively, we measured the histogram intersection of each shared covariate's value distribution between the two regions, on a common 100-bin range. The mean overlap is $0.63$, ranging from $0.47$ for bio5x5n to $0.77$ for bio3x3n, with slope at $0.56$ and aspect at $0.71$. The regions are therefore similar in the shape of their covariate distributions but far from identical: about a third of the distribution mass does not overlap. Pedriza is also steeper on average (mean slope $30.5$ against $22.2$). The models were first applied to Pedriza \emph{zero-shot}, with no retraining. Table~\ref{tab:pedriza} reports that comparison, together with the effect of fine-tuning on Pedriza itself.

\begin{figure}[ht]
\centering
\includegraphics[width=1\textwidth]{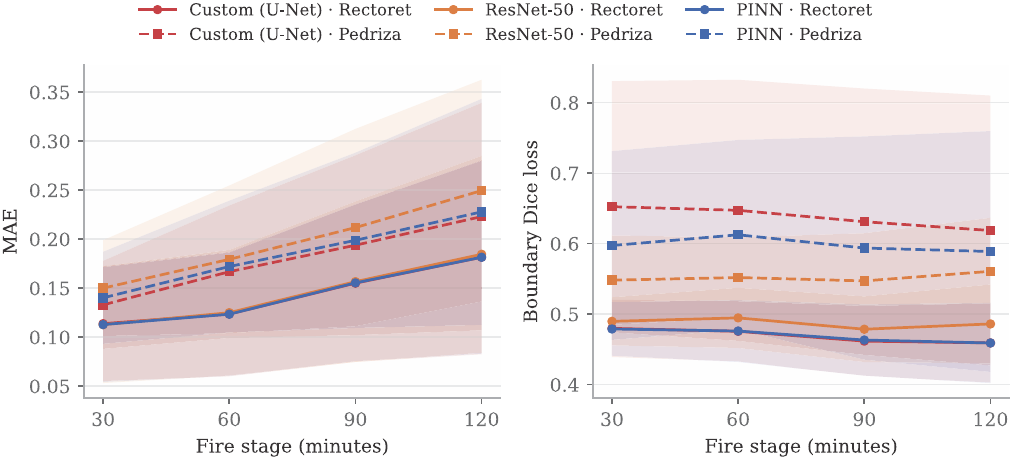}
\caption{Cross-region performance against fire stage, per model, evaluated zero-shot. Left: MAE. Right: boundary Dice loss. Solid lines are Rectoret, dashed lines Pedriza. Shaded bands are $\pm1$ standard deviation across evaluation simulations. Swin-Unet is absent from the region logs.}
\label{fig:pedriza-gen}
\end{figure}

At the model level (Fig.~\ref{fig:pedriza-gen}), the trained surrogates were evaluated on Pedriza against their Rectoret performance across fire stage (40--120 min), using an MCC-based score and a boundary Dice-loss-based overlap error. Both metrics show the same pattern. The Pedriza curves follow the shape of the Rectoret curves, offset by a roughly constant amount rather than collapsing. Table~\ref{tab:pedriza} gives the per-model figures. Zero-shot Dice loss rises from $0.469$--$0.487$ on Rectoret to $0.552$--$0.637$ on Pedriza. Fine-tuning on Pedriza recovers most of that gap for Custom and PINN, both reaching $0.481$ after 50 epochs. ResNet-50 behaves differently. It improves only up to epoch~2 and then degrades slowly to $0.519$. It is also the model that transfers best zero-shot, even though it is the weakest of the three on Rectoret.

\begin{table}[ht]
\centering
\small
\caption{Cross-region comparison, averaged over fire stages 30--120~min, with $\pm1$ standard deviation across evaluation simulations. Zero-shot means the Rectoret-trained model applied to Pedriza without retraining. The last row is Pedriza after 50 fine-tuning epochs on Pedriza.}
\label{tab:pedriza}
\begin{tabular}{llccc}
\toprule
Region & Metric & Custom & ResNet-50 & PINN \\
\midrule
\multirow{2}{*}{Rectoret} & MAE  & $0.143\pm0.075$ & $0.145\pm0.077$ & $0.143\pm0.075$ \\
 & Dice & $0.469\pm0.047$ & $0.487\pm0.045$ & $0.469\pm0.047$ \\
\midrule
\multirow{2}{*}{Pedriza (zero-shot)} & MAE  & $0.179\pm0.080$ & $0.198\pm0.085$ & $0.185\pm0.080$ \\
 & Dice & $0.637\pm0.186$ & $\mathbf{0.552}\pm0.066$ & $0.598\pm0.149$ \\
\midrule
Pedriza (fine-tuned) & Dice & $\mathbf{0.481}\pm0.055$ & $0.519\pm0.061$ & $\mathbf{0.481}\pm0.053$ \\
\bottomrule
\end{tabular}
\end{table} We report this as a descriptive comparison only. Establishing that the models generalise would require a second region mapped at the same 2~m resolution, so that scale and region are not varied together, and further regions beyond that.

Other cross-region transfer studies report the same pattern: a roughly constant offset instead of a collapse. A cross-county wildfire-risk study found that transfer depends sharply on ecological similarity \cite{liu2025wildfiregenome}. Transfer was near chance between dissimilar counties and strong between similar ones. The Rectoret--Pedriza offset should therefore be read against how similar the two regions are, which we have not yet quantified. An attention-based ConvLSTM study found that pairwise self-attention transferred poorly across regions, while patchwise, window-local self-attention transferred better \cite{masrur2024convlstm}. This bears directly on Swin-Unet, whose attention is also window-based and connects back to the interpretability results above.

\section{Discussion}
\label{sec:discussion}

The main finding is that surface fuel load alone accounts for most of the explainable variance in burn probability at the pixel level. The vegetation-continuity descriptors add little, because they largely repeat one another. In practice, this suggests that fuel management should focus specifically on surface fuel loading rather than the entire suite of continuity metrics. This result requires one important caveat. The surface fuel load raster comes from the same BEHAVE-Anderson fuel classification that the simulator uses for fire spread. Therefore, it is not independent of the process that generated our target variable and some of its apparent dominance may reflect this common source rather than a purely physical effect. Confirming this finding with an independently measured fuel load product is a natural next test.

The wind seemed inconsistent at first. It has a strong impact at the simulation level, but was rated low in the channel-specific importance analysis. The rotation experiment solves this problem. The models use wind direction and the previous low ranking was partly an artefact of encoding wind as a constant across the area, which weakens methods that look for spatial variability. The physical effect is real, but secondary: wind reversal shifts the forecast by about a fifth compared to removing the distance from the fire front.

On generalisation, two further directions from the literature are relevant. First, our models give a single point prediction. A recent denoising-diffusion surrogate for a stochastic CA burn-probability simulator samples an ensemble instead \cite{yu2026diffusion}. It beats a deterministic model of the same architecture on accuracy, spatial coherence and distributional quality at once. Our dataset is itself a stochastic ensemble, so a probabilistic surrogate is a natural next step. Second, work on differentiable-Eikonal fire spread treats cross-region transfer as a test of whether the learned covariate--physics relationships are \emph{universal}. It trains on 11 wildfire scenes and tests on 4 held-out scenes that never appear in training \cite{gahtan2026eikonal}. That protocol is stricter than our single train/test region split and worth adopting if more regions become available. Two limitations remain. The training data is completely synthetic, without validation against observed fires and covers one type of fuel.

\section{Conclusions and Future Work}
\label{sec:conclusion}

This paper compares four deep learning models for predicting wildfire spread in an ensemble of high-resolution simulations: a custom U-Net, a ResNet-50 network with transfer learning, a physics-informed model and a Swin-Unet transformer. This is combined with exploratory analysis and interpretability. Three results stand out. First, surface fuel load is the single strongest driver of burn probability, well ahead of all other covariates. Second, there is a clear separation between the architectures: convolutional models rely primarily on distance from the fire front, while Swin-Unet assigns more weight to terrain and fuel, consistent with results obtained in an unrelated dataset. Third, the comparison with Pedriza shows a consistent shift, not a decrease in accuracy. This is encouraging for application beyond a single region, although it does not yet prove model transferability. Two further steps must then be taken. The first is to test the surface fuel load against an independently measured fuel product, as the raster used here is from the same classification used by the simulator. The second is to further analyse the 2~m resolution region, which would allow us to separate the area change from the scale change, which we cannot do in the current comparison.


\bibliographystyle{splncs04}
\bibliography{references}

\end{document}